\documentclass[letterpaper, 10 pt, conference]{ieeeconf}  % Comment this line out if you need a4paper

\IEEEoverridecommandlockouts                              % This command is only needed if 
\usepackage{amsmath} % assumes amsmath package installed
\usepackage{subfigure}
\usepackage{float}
\usepackage{todonotes}
\usepackage{amsmath}
\usepackage{amssymb}
\usepackage{diagbox}
\usepackage{makecell}
\usepackage{amsmath}
\usepackage{url}
\usepackage{hyperref}
\usepackage{bm}
\usepackage{array}
\let\labelindent\relax
\usepackage{enumitem}
\usepackage{color,soul}
\usepackage{leftindex}
\usepackage{graphicx}
\usepackage{tensor}   % For tensor notation with better spacing
\usepackage{stackengine}

\usepackage{adjustbox}
\usepackage{threeparttable}
\usepackage{multirow}
\usepackage{booktabs}

\title{\LARGE \bf
Aerial Manipulation with Contact-Aware Onboard Perception and Hybrid Control  %Contact-Constrained VIO and Visual Servoing for Robust Aerial Manipulation
}

\author{Yuanzhu Zhan$^{*1}$, Yufei Jiang$^{*1}$ Muqing Cao$^{2}$, Junyi Geng$^{1}$
\thanks{$^{*}$ Equal contribution.}
\thanks{$^{1}$ Department of Aerospace Engineering, Pennsylvania State University, University Park, PA, 16802, USA. 
{\tt\footnotesize \{yvz6008, yufei\_jiang, jgeng\}@psu.edu}}%
\thanks{$^{2}$ Robotics Institute, Carnegie Mellon University, Pittsburgh PA 15213, USA.
{\tt\footnotesize muqingc@andrew.cmu.edu}}%}% 
}

\begin{document}

\maketitle
\thispagestyle{empty}
\pagestyle{empty}

%%%%%%%%%%%%%%%%%%%%%%%%%%%%%%%%%%%%%%%%%%%%%%%%%%%%%%%%%%%%%%%%%%%%%%%%%%%%%%%%
\begin{abstract}

Aerial manipulation (AM) promises to move Unmanned Aerial Vehicles (UAVs) beyond passive inspection to contact-rich tasks such as grasping, assembly, and in-situ maintenance. Most prior AM demonstrations rely on external motion capture (MoCap) and emphasize position control for coarse interactions, limiting deployability. We present a fully onboard perception–control pipeline for contact-rich AM that achieves accurate motion tracking and regulated contact wrenches without MoCap. The main components are (1) an augmented visual–inertial odometry (VIO) estimator with contact-consistency factors that activate only during interaction, tightening uncertainty around the contact frame and reducing drift, and (2) image-based visual servoing (IBVS) to mitigate perception–control coupling, together with a hybrid force–motion controller that regulates contact wrenches and lateral motion for stable contact. Experiments show that our approach closes the perception-to-wrench loop using only onboard sensing, yielding an velocity estimation improvement of 66.01\% at contact, reliable target approach, and stable force holding—pointing toward deployable, in-the-wild aerial manipulation.

%Accurate state estimation is critical for aerial robots operating in GPS-denied environments.  External motion capture (MoCap) system comes to a rescue to overcome these limitations, but such infrastructure is costly, restricted, and impractical for real-world deployment. Visual-Inertial Odometry (VIO) has become a standard solution, but it inevitably suffers from drift and degraded accuracy during long-term missions. This paper presents a novel framework that eliminates the need for MoCap by combining our innovative contact-constrained VIO with visual servoing. A lightweight rod mounted on the fully-actuated hexarotor interacts with the environemnt, and when contact occurs with a vertical wall, the resulting geometric constraints are incorporated into the VIO estimation process. Furthermore, to bypass the reliance of accurate position estimation, visual servoing is more suitable for our specific scenario, which only need the velocity estimation as feedback. A drone shooting scenario is designed in this paper, experimental results show that the drone can reach the operation point accurately under the guidance of visual servoing, and perform precise force control for inspection purpose. The statistical results imply that it is so promising to transfer our system to outdoor opoeration scenarios in the near future.

\end{abstract}

%%%%%%%%%%%%%%%%%%%%%%%%%%%%%%%%%%%%%%%%%%%%%%%%%%%%%%%%%%%%%%%%%%%%%%%%%%%%%%%%
\section{Introduction}
\label{sec: intro}
% Aerial manipulation is important, overcome purly passive UAV. The challenges are XXX
Despite the rapid growth, most Unmanned Aerial Vehicles (UAVs) today are limited to passive tasks like visual inspection, surveillance, and remote sensing~\cite{hu2023off, qiu2025airio}, where the UAVs avoid collision without physical interaction. Aerial manipulation (AM), where UAVs equipped with manipulators performing contact-based tasks such as grasping, assembly, transporting, and manipulating objects at height, will unlock transformative applications like infrastructure maintenance (e.g., bridge repair and relic restoration), precision agriculture, and contact inspection in manufacturing. Such capabilities can reduce human risk, enhance efficiency in labor-intensive sectors, and expand access to extreme or hazardous environments. The key challenge lies in the dynamical coupling between the UAVs and their interactions with other entities, where floating base body motion affects the end-effector and vice versa, as well as the inherent instability of multirotor platforms.

% Most of exisiting aerial manipulation work focus on indoor from mechanical design, controllers design, ...on various kind of tasks xxx. There are few outdoor demos. 
Most aerial manipulation demonstrations to date have been conducted in controlled lab settings—e.g., contact-based inspection~\cite{bodie1905omnidirectional, guo2024aerial}, door opening~\cite{lee2020aerial}, valve turning~\cite{brunner2022planning}, cleaning~\cite{alkaddour2023novel}, and letter drawing~\cite{guo2024flying}. These studies typically rely on external motion capture (MoCap) to provide near–ground-truth localization and state estimates. That assumption breaks down in real-world deployments, where external tracking is unavailable and GPS can be unreliable or denied especially near large infrastructure, indicating the necessity of fully onboard perception and control.

\begin{figure}
    \centering
    \includegraphics[width = 1\linewidth]{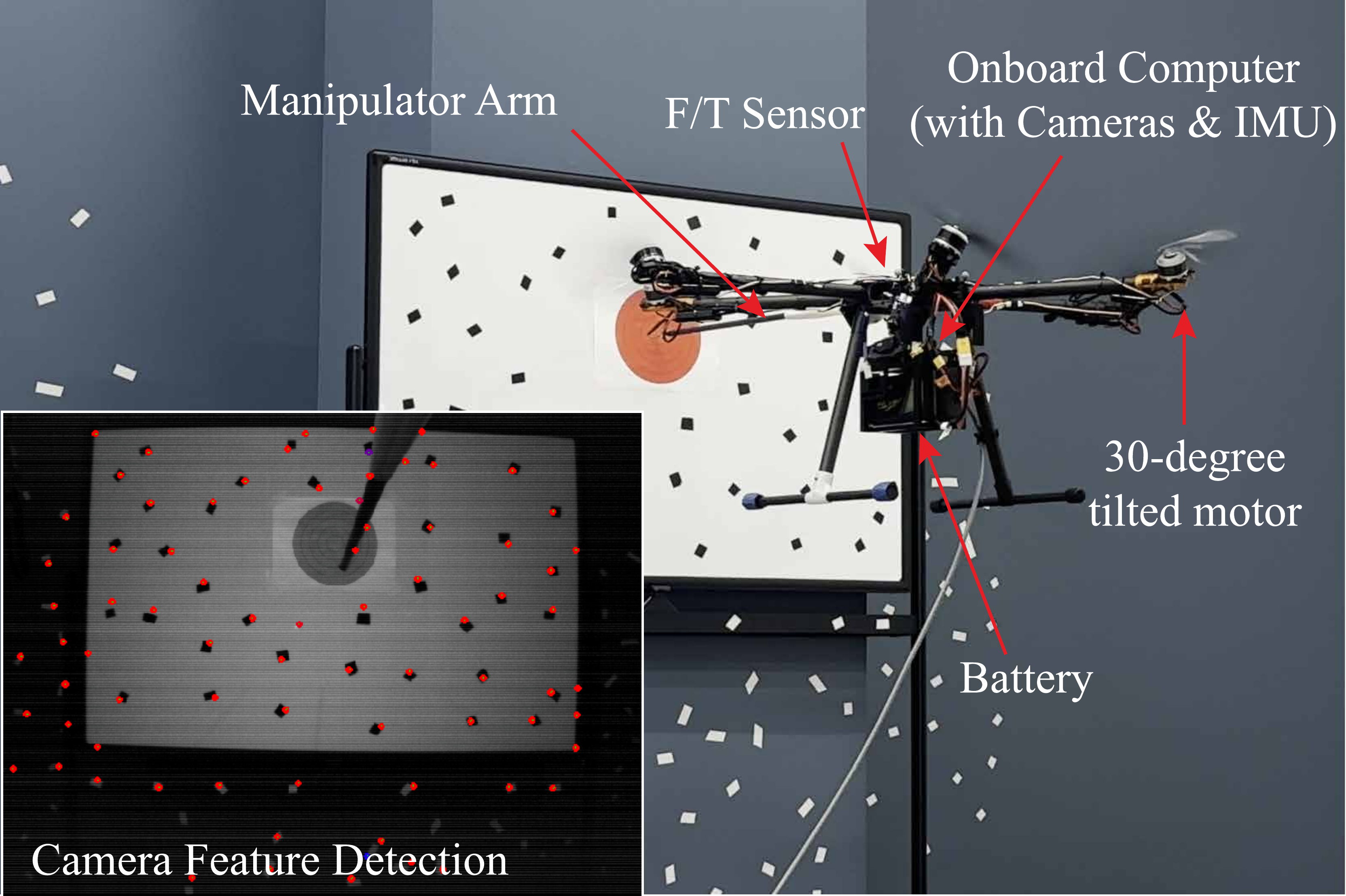}
    \caption{Our fully-actuated aerial manipulator performs contact inspection using solely onboard sensing for motion tracking and force regulation. The inset shows the camera-based feature detection of our contact-aware visual inertial odometry.} %Experimental platform: a fully actuated hexarotor equipped with a manipulator arm and a force/torque (F/T) sensor for contact interaction. The onboard computer integrates cameras and an IMU for visual–inertial state estimation, while the battery powers the system. Each motor is tilted by 30 degrees to enable full actuation in wrench space. }
    \label{fig: CoverPage}
    \vspace{-1.5em}
\end{figure}

% Although there are some work starting to research on aerial manipulation in the wild --> outdoor demos, they dealt with coarse pick and place tasks relying on specific mechanical design without considering contact and target force. only pure position 
Emerging efforts have begun to investigate aerial manipulation in the wild with outdoor demonstrations, including GPS/RTK-assisted sensor retrieval~\cite{hamaza2020design,wen2025full}, total-station-based inspection~\cite{jimenez2019contact}, and vision-based coarse pick-and-place of large objects~\cite{bauer2024open}. However, these systems typically address simple or low-precision tasks dominated by position control, or rely on unconventional mechanical designs, e.g., soft grippers~\cite{ubellacker2024high} and compliant continuum manipulators~\cite{peng2025dexterous} for position-based grasping.  
To our knowledge, they rarely model or regulate contact wrenches (force/torque), which is critical for contact-rich aerial manipulation tasks such as aerial contact inspection and peg-in-hole for maintenance, where maintaining specified wrenches is essential.

We aim to achieve precise aerial manipulation tasks in real world using only onboard sensing for maintaining accurate motion tracking and regulated contact wrenches. This is challenging for several reasons. First, the localization accuracy of standard vision-based Simultaneous Localization and Mapping (SLAM) / Visual Inertial Odometry (VIO) is often insufficient for simultaneous motion and force/torque control, especially near infrastructure where features are sparse or occluded. In addition, due to the tight dynamic coupling between the UAV body and the manipulator, perception and control are deeply intertwined: estimation errors can destabilize interaction control, while suboptimal control further degrades perception. These factors motivate a jointly designed perception–control framework.

% To address these gaps, we propose a contact aware perception and control framework for robust aerial manipulation. 
To address these challenges, this paper develops a contact-aware, fully onboard perception–control pipeline for aerial manipulation. Specifically, we augment the state-of-the-art VIO algorithm~\cite{qin2017vins, qin2018online, qin2019a, qin2019b} with contact consistency factors that activate only during interaction to stabilize estimates at contact. This tightens the uncertainty around the contact frame and reducing drift that would otherwise degrade interaction quality.
To mitigate perception–control coupling issue of aerial manipulation, we adopt visual servoing to co-design perception and control. Image-based visual servoing (IBVS) directly uses image-space feedback to compute commands and thus does not require full UAV pose estimation~\cite{chaumette2006visual}. We thus use only the VIO-provided body velocity for control feedback. Finally, a hybrid force–motion controller regulates contact wrenches along the surface normal while tracking lateral motion, maintaining both target pose and desired wrench during manipulation.
In summary, the main contributions of
this work are:
\begin{itemize}[leftmargin=*]
\item We propose a contact-aware, \textit{onboard} perception–control pipeline for contact-rich aerial manipulation that achieves both precise motion tracking and wrench regulation \textit{without external motion capture}.
\item We augment factor-graph VIO with \textit{contact-consistency} constraints that activate only during interaction to improve estimation performance at contact.
\item We develop an image-based visual servoing scheme that couples perception and control, together with a hybrid force–motion controller that regulates contact wrenches and lateral motion.
\item We present simulation and real world experiments to demonstrate the effectiveness of accurate motion tracking and stable force holding under visual guidance.
\end{itemize}

\section{Related Works}
\label{sec: related works}

\subsection{Aerial Manipulation}
Research on aerial manipulation spans diverse tasks, such as aerial grasping~\cite{mellinger2011design}, aerial docking~\cite{choi2022automated}, drilling~\cite{ding2021design}, screwing~\cite{schuster2022automated}, and pushing a target~\cite{lee2020aerial1}. However, most demonstrations rely on external motion-capture (MoCap) for localization and are conducted indoors. Prior work has largely emphasized mechanical design~\cite{lanegger2022aerial, hamaza2020design, zhao2018design} and dynamic analysis/control~\cite{wang2023millimeter, dong2021centimeter, cuniato2022power}, with comparatively less attention to onboard perception~\cite{blochlinger2025reliable}. 
Although emerging efforts push aerial manipulation into the wild with outdoor demonstrations, they often depend on external infrastructure such as GPS/RTK~\cite{hamaza2020design,wen2025full} or total stations~\cite{jimenez2019contact}, and only a few use onboard sensing with standard VIO/SLAM for coarse pick-and-place of large objects~\cite{blochlinger2025reliable}. These typically employ position-only control without explicit wrench regulation during physical interaction. In contrast, our goal is precise, MoCap-free aerial manipulation using only onboard sensing to control both motion and wrench.

\subsection{Onboard Perception for Aerial Robots}
Onboard perception for UAVs is well studied across platforms and tasks ranging from inertial+GPS navigation~\cite{hoffmann2007quadrotor} to LiDAR/sonar mapping~\cite{zhou2021fuel} and, more recently, lightweight vision pipelines for tracking~\cite{ghosh2023airtrack}, landing~\cite{lin2017monocular}, and inspection~\cite{xing2023autonomous}. In aerial manipulation, however, onboard perception remains comparatively less mature. Existing aerial manipulation systems often employ VIO/SLAM (sometimes aided by fiducial markers) for approach behaviors such as grasping~\cite{ubellacker2024high}, perching~\cite{mao2021aggressive}, or package delivery~\cite{li2021cooperative}. However, contact is typically treated as a disturbance where texture loss, self-occlusion, and low-speed motion weaken visual constraints when accuracy is most critical. Consequently, MoCap-free demonstrations tend to emphasize coarse placement or light interaction and rely on position-only control without explicit wrench regulation~\cite{blochlinger2025reliable, ubellacker2024high}. Outside the aerial domain, contact-aided estimation has improved observability by injecting no-slip/contact constraints into factor-graph formulations for legged or mobile manipulation~\cite{hartley2018legged, buchanan2024online}. Inspired by this principle, we integrate contact-consistency into a VIO that activates only during interaction to stabilize pose/velocity at contact. To connect perception to action, visual servoing provides an interface that has been widely used for UAV tasks~\cite{xu2022image, byun2024image}. Yet, prior aerial manipulation pipelines rarely close the full perception-to-wrench loop without MoCap; those that do are typically validated only in simulation~\cite{he2023image}. In contrast, our work explicitly bridges targeting and contact using only onboard sensing leveraging velocity estimates from our contact-aware VIO to drive visual servoing and hybrid force–motion control.

%; those that do are typically validated only in simulation~\cite{he2023image}. In contrast, we bridge targeting and contact using only onboard sensing: velocity estimates from our contact-aware VIO drive visual servoing for approach and a hybrid force–motion controller for interaction.

\section{System Overview}
\label{sec: SystemOverview}

\subsection{System Design}
\begin{figure*}
    \centering
    \includegraphics[width = 0.95\linewidth]{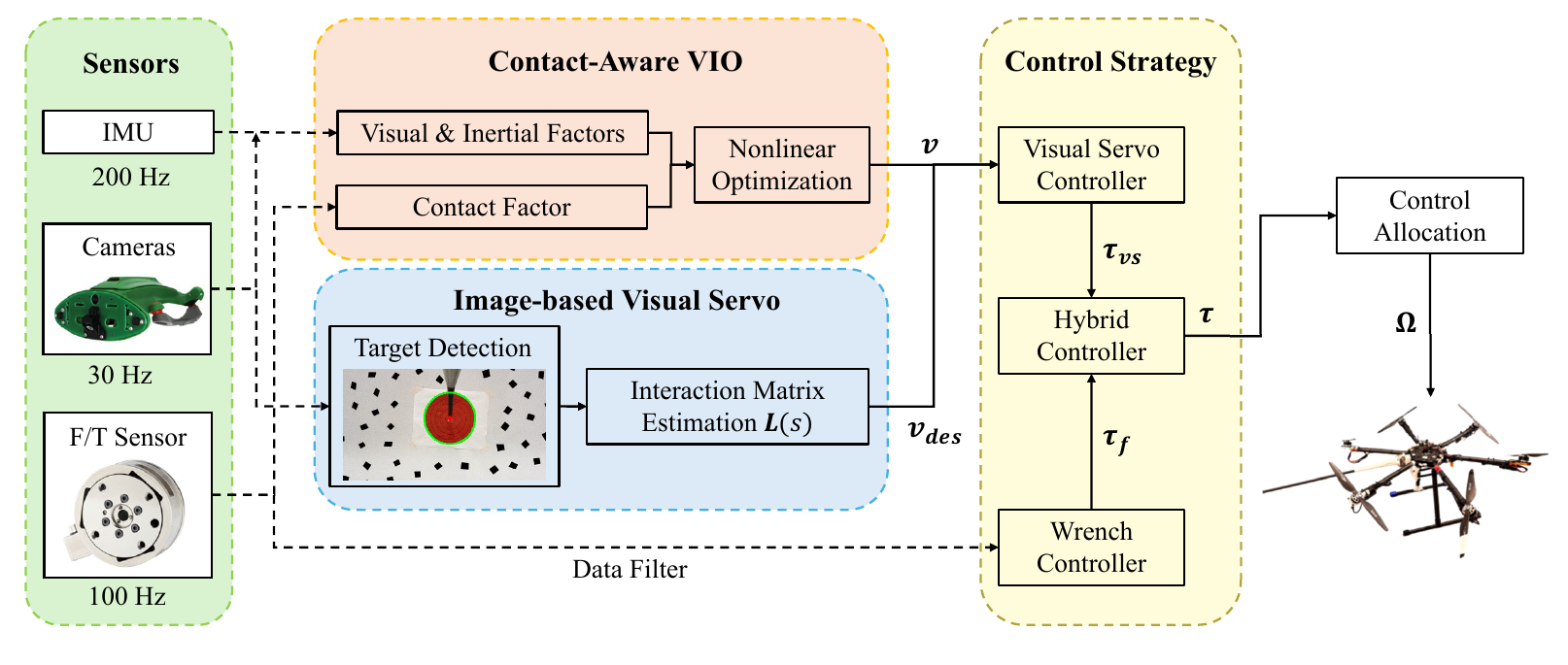}
    \caption{Overview of the proposed pipeline. A contact-aware VIO module fuses 
visual, inertial, and force/torque measurements to provide state estimates and 
activates contact factors during interaction. An image-based visual servo (IBVS) 
drives the vehicle during the approach phase using VIO-estimated body velocity. 
Upon contact, a hybrid motion–wrench control strategy regulates contact forces 
and tracks lateral motion. Finally, a control allocation module maps the commanded 
wrenches into rotor speeds $\boldsymbol{\Omega}$ for execution on the fully-actuated hexarotor platform.}
    \label{fig: pipeline}
    \vspace{-2em}
\end{figure*}

\label{subsec: systemDesign}
As illustrated in Figure \ref{fig: CoverPage}, our customized aerial platform is a hexarotor (Tarot T960) with six rotors, each tilted by 30 degrees in alternating left and right directions. Every arm carries a brushless motor driving a 12-inch propeller. Leveraging its fully actuated nature, we mount a zero degree-of-freedom (DOF) manipulator arm to the front of the vehicle without the need for additional complex mechanical components. A 6-axis force/torque (F/T) sensor ATI Gamma is installed at the base of the manipulator for interaction forces and moments measurement.

As for the avionics, we use the compact ModalAI VOXL 2 Flight Deck\footnote{\url{https://www.modalai.com/}}, which integrates an autopilot running PX4 firmware, an onboard computer, an inertial measurement unit (IMU), a barometer, and multiple optical sensors into a single unit. The platform provides a stereo pair for VIO and an RGB camera for visual servoing, meeting our sensing requirements.

\subsection{Notations}
We define five reference frames: the world inertial frame $\mathcal{W}$, the vehicle body frame $\mathcal{B}$, the camera frame $\mathcal{O}$, the F/T sensor frame $\mathcal{S}$, and the contact frame $\mathcal{C}$. 

The inertial frame $\mathcal{W}$ is a north-east-down frame whose origin is located at the UAV's takeoff position. The body frame is denoted as $\mathcal{B} = \{O_\mathcal{B}, \hat{x}_b, \hat{y}_b, \hat{z}_b\}$, where $O_\mathcal{B}$ represents the vehicle's center of mass, and $\hat{x}_b, \hat{y}_b, \hat{z}_b$ are unit vectors pointing forward-left-up relative to the vehicle, respectively. The camera frame is defined as $\mathcal{O} = \{O_\mathcal{O}, \hat{x}_o, \hat{y}_o, \hat{z}_o\}$, where $O_\mathcal{O}$ is the optical center of the camera, and $\hat{x}_o, \hat{y}_o, \hat{z}_o$ are unit vectors pointing right-down-forward relative to the camera. The F/T sensor frame is defined as $\mathcal{S} = \{O_\mathcal{S}, \hat{x}_s, \hat{y}_s, \hat{z}_s \}$, whose $O_\mathcal{S}$ is at the base of the manipulator where the F/T sensor is mounted, and $\hat{x}_s, \hat{y}_s, \hat{z}_s$ are unit vectors aligned with the left, down, and up (left-handed coordinates) directions of the sensor. The contact frame is given by $\mathcal{C} = \{O_\mathcal{C}, \hat{x}_c, \hat{y}_c, \hat{z}_c\}$, where $O_\mathcal{C}$ corresponds to the end-effector tip at the contact point, and $\hat{x}_c, \hat{y}_c, \hat{z}_c$ are unit vectors aligned with the surface normal (pointing inward), left, and up directions of the contact surface.

Furthermore, $\mathbf{R}_a^b \in \text{SO}(3)$ and $\mathbf{T}_a^b \in \text{SE}(3)$ define the rotation and transformation matrix for vector from frame $b$ to frame $a$. The right superscripts, e.g., $\mathbf{v}^\mathcal{W}$, indicates that the vector $\mathbf{v}$ is expressed in frame $\mathcal{W}$. $\mathbf{0}_{a\times b}$ denotes a zero vector of dimension $a \times b$.

\subsection{System Dynamics}
Our platform is a fully-actuated aerial manipulator. Unlike conventional coplanar multirotor whose thrust is fixed along the body z-axis and must tilt the entire vehicle to generate lateral forces, a fully-actuated design can produce a full 6 DOF wrench and thus regulate translation and orientation independently. This decoupling is well suited to contact-rich tasks such as aerial contact inspection.
%A fully-actuated hexarotor UAV serves as the base of our aerial manipulation system. Thanks to its capability of generating six-dimensional force and torque independently, it could perform flexible lateral motion without attitude change, which provide tremendous convenience for our precise aerial operations.

The dynamics of our hexarotor UAV can be modeled using Euler-Lagrangian method as following:
\begin{equation}
    \mathbf{M}\dot{\bm{v}} + \mathbf{C}\bm{v} + \mathbf{G} = \boldsymbol{\tau} + \boldsymbol{\tau}_{ext}
\end{equation}
with inertia matrix $\mathbf{M} \in \mathbb{R}^{6\times6}$, the centrifugal and Coriolis matrix $\mathbf{C} \in \mathbb{R}^{6\times6}$, the gravity vector $\mathbf{G} \in \mathbb{R}^{6}$. $\bm{v} = [\mathbf{v}^\top, \bm{\omega}^\top]^\top$ and $\dot{\bm{v}}$ represent the system twist and acceleration respectively. $\boldsymbol{\tau} \in \mathbb{R}^{6}$ is the control wrench from the actuators, while $\boldsymbol{\tau}_{ext} \in \mathbb{R}^{6}$ is other external wrench (e.g. wind or contact). %Here, we consider $\boldsymbol{\tau}_{ext}$ as the contact wrench.

\subsection{Method Overview}
\label{subsec: methodOverview}
We target aerial manipulation tasks that require precise motion tracking and regulated contact wrenches. The representative scenario is approaching a surface and sustaining a prescribed normal force, such as in aerial contact inspection or peg-in-hole maintenance. We focus on the problem to the approach and interaction phases because long-range, free-flight guidance can be handled by other standard and mature methods and is outside the scope of this work. The proposed pipeline as shown in Figure.~\ref{fig: pipeline} has three key components: (1) A contact-aware VIO processes stereo imagery to produce vehicle/end-effector state estimates, activating contact factors only during interaction. (2) During approach, image-based visual servoing (IBVS) drives the platform using only the VIO-estimated body velocity. (3) Upon contact, a hybrid force–motion controller regulates the normal wrench while IBVS continues to track lateral motion, closing the perception-to-wrench loop with onboard sensing.

%We focus on the aerial manipulation tasks that require accurate motion tracking and regulated contact wrenches. Here, we consider a scenario that the AM needs to fly towards the target and hold reference force, like the aerial contact inspection or peg-in-hole for aerial maintenance.  We make an assumption that we only focus on the flight phase when the AM is approaching to the target of interest and then perform physical interaction because other mature guidance methods exist when the AM is far away from the infrastructure and conducts free flight. The proposed system pipeline is shown in Figure~\ref{}. First, a contact aware VIO algorithm is developed which takes into the stereo image pair and output the state estimates of the aerial manipulator. Second, an image-based visual servo guides the AM to the target in the approaching phase leveraging only the velocity estimates from the contact aware VIO. Finally, when the end-effector of the AM touches the target, a hybrid force and motion control activates and keeps running to maintain the target force while the IBVS help to regulate the target motion. 

\section{Contact-aware Visual-inertial Odometry}
\label{sec: VIO}
\begin{figure}
    \centering
    \includegraphics[width=0.49\textwidth]{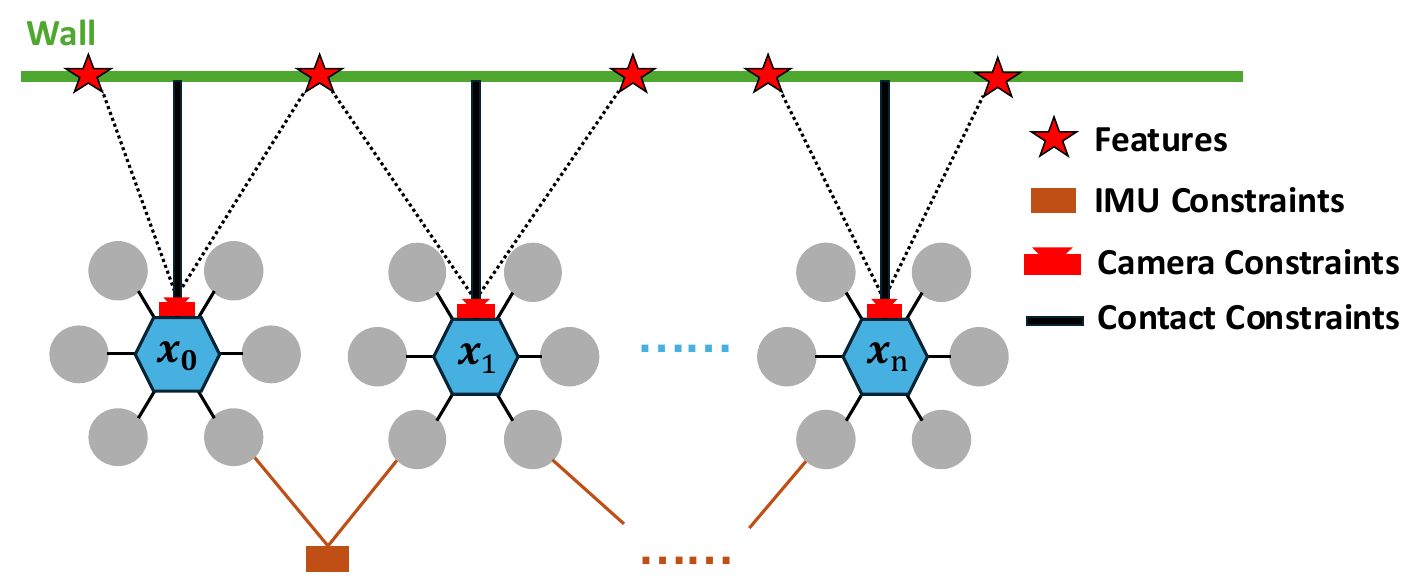}
    \vspace{-2em}
    \caption{Pose graph of the proposed contact-aware VIO}
    \label{fig:pose_graph}
    \vspace{-3mm}
\end{figure}

Our method builds upon VINS-Fusion \cite{qin2017vins, qin2018online}, where visual–inertial state estimation is expressed within a factor graph framework, and the system state is recovered through joint optimization of camera reprojection errors and IMU pre-integration constraints. While this formulation achieves high accuracy in general scenarios, it may suffer from drift when external interactions occur. To mitigate this issue, we extend it by introducing contact factors that leverage robot–environment interactions. In particular, when the system is in contact with a wall, the relative motion at the contact point is constrained, and this information is encoded as an additional factor in the graph as shown in Figure \ref{fig:pose_graph}. By incorporating such constraints, the estimator gains improved robustness and reduced drift, especially in environments with limited visual features or long-term operation.

\subsection{Visual-inertial Odometry}
We formulate visual–inertial odometry with contact constraints as a nonlinear optimization problem over a sliding window of states. 
The full state vector is defined as
\begin{equation}
    \boldsymbol{\mathcal{X}} =
    \left[ \bm{x}_0, \bm{x}_1, \ldots, \bm{x}_n, 
    \bm{x}_c^\mathcal{B}, \gamma_0, \gamma_1, \ldots, \gamma_m \right],
\end{equation}
where each aerial manipulator state $\mathbf{x}_k$ at time $k \in [0,n]$ includes the position, velocity, orientation, and bias parameters of the accelerometer and gyroscope, 
and we assume that the body frame of the aerial manipulator is aligned with the IMU frame.

\begin{equation}
    \bm{x}_k =
    \left[ 
    \bm{p}_k^\mathcal{W}, \;
    \mathbf{v}_k^\mathcal{W}, \;
    \bm{q}_k^\mathcal{W}, \;
    \bm{b}_a, \;
    \bm{b}_g
    \right]
\end{equation}

and $\bm{x}_c^\mathcal{B} = \left[ \bm{p}_c^\mathcal{B}, \bm{q}_c^\mathcal{B} \right]$ denotes the translation and orientation of camera in the body frame. 
The variables $\gamma_i$ represent the inverse depths of visual landmarks.

Given this state representation, the state estimation problem is expressed as minimizing the sum of residuals from different sensing modalities:
\begin{equation}
    \min_{\boldsymbol{\mathcal{X}}} \;
    \| r_p - H_p \boldsymbol{\mathcal{X}} \|^2 + 
    \sum_{(i,j) \in \mathcal{C}} 
    \big\| r_{\text{visual}}^{ij}(\boldsymbol{\mathcal{X}}) \big\|^2 
    + \sum_{k \in \mathcal{I}} 
    \big\| r_{\text{IMU}}^{k}(\boldsymbol{\mathcal{X}}) \big\|^2
   .
\end{equation}

$r_{\text{visual}}^{ij}$ denotes the reprojection residual of feature $j$ observed at time $i$, $r_{\text{IMU}}^k$ denotes the IMU pre-integration residual between consecutive states. 
This joint optimization ensures consistency between visual and inertial measurements and provides accurate state estimation in general scenarios. 
The visual residuals and IMU residuals follow the standard formulation in \cite{qin2017vins}, and we refer readers to that paper for more detailed derivations.

\subsection{Contact Factor}
To further enhance robustness in scenarios involving robot–environment interactions, we introduce a contact factor into the factor graph. 
When the system establishes contact with a wall, the relative motion at the contact point can be assumed to follow physical constraints, which are encoded as additional residuals in the optimization. 
Formally, the extended objective becomes
\begin{align}
    \min_{\boldsymbol{\mathcal{X}}} ~
    &\| r_p - H_p \boldsymbol{\mathcal{X}} \|^2 + 
    \sum_{(i,j) \in \mathcal{V}} \| r_{\text{visual}}^{ij}(\boldsymbol{\mathcal{X}}) \|^2 
    + \sum_{k \in \mathcal{I}} \| r_{\text{IMU}}^{k}(\boldsymbol{\mathcal{X}}) \|^2 \notag\\
    &+ \sum_{k \in \mathcal{F}} \| r_{\text{contact}}^{k}(\boldsymbol{\mathcal{X}}) \|^2,
\end{align}
where $r_{\text{contact}}^{k}$ enforces the constraint induced by wall contact at time $k$. 
Here, $\mathcal{V}$ denotes the set of all visual feature observations, 
$\mathcal{I}$ denotes the set of all IMU measurement intervals, 
and $\mathcal{F}$ denotes the set of all contact events.

The contact factor reduces drift and improves estimation accuracy by anchoring the states to the environment during contact events, especially under feature-sparse or long-term operation.

Concretely, the contact event is triggered by a force threshold. After we switch to force holding mode, the robot is constrained along the wall normal direction. 
Between two consecutive states, the signed distance to the wall is preserved, and the velocity projected onto the wall normal is zero. 
These constraints are expressed as
\begin{equation}
r_{\text{contact}}^{k}(\bm{x}) =
\begin{bmatrix}
\bm{n}^\top \big(\bm{p}_{k+1}^\mathcal{W} - \bm{p}_{k}^\mathcal{W}\big) \\[6pt]
\bm{n}^\top \mathbf{v}_{k}^\mathcal{W}
\end{bmatrix},
\end{equation}
where $\bm{n}$ denotes the wall normal, $\bm{p}_{k}$ and $\bm{p}_{k+1}$ are positions at consecutive time steps, and $\bm{v}_{k}$ is the velocity at time $k$ in the world frame. 
By jointly constraining both position and velocity along the contact normal, the factor further stabilizes the state estimation in challenging environments.

% To further improve robustness, the covariance of the contact factor is adaptively adjusted according to the force sensor readings. 
% Intuitively, when the contact force is stable and large, the system is confident that the robot is in firm contact with the environment, and the corresponding constraint should be enforced more strongly. 
% Conversely, if the force sensor signal is noisy or unstable, this indicates that the contact is not reliable, and the constraint should be relaxed. 
% Formally, the information matrix of the contact residual is defined as
% \begin{equation}
%     \mathbf{P}_{\text{contact}} = \alpha \, \mathrm{Var}(F),
% \end{equation}
% where $\mathrm{Var}(F)$ denotes the variance of the measured contact force along the constrained direction, and $\alpha$ is a scaling coefficient. The variance is computed over a sliding window of recent force sensor data, which captures the stability of the contact interaction. 
% This formulation ensures that the influence of the contact factor is automatically weakened when the force measurements are unreliable, thereby preventing the introduction of spurious constraints.

% This adaptive weighting ensures that the contact factor provides useful information only when a reliable contact is detected, thereby avoiding the introduction of spurious constraints when the robot is not strictly in contact with the wall.

To further improve robustness, the covariance of the contact factor is adaptively adjusted according to the force sensor readings. 
Intuitively, when the contact force is stable and large, the system is confident that the robot is in firm contact with the environment, and the corresponding constraint should be enforced more strongly. 
Conversely, if the force sensor signal is noisy or unstable, this indicates that the contact is not reliable, and the constraint should be relaxed. 
Formally, the information matrix of the contact residual is defined as
% \begin{equation}
%     \mathbf{P}_{\text{contact}} = \alpha \, \mathrm{Var}(F),
% \end{equation}
% where $\mathrm{Var}(F)$ denotes the variance of the measured contact force along the constrained direction, and $\alpha$ is a scaling coefficient. 
% The variance is computed over a sliding window of recent force sensor data, which captures the stability of the contact interaction. 
\begin{equation}
    \mathbf{P}_{\text{contact}} 
    = \alpha \, \frac{1}{N} \sum_{i=1}^{N} \Big(F_i - \tfrac{1}{N}\sum_{j=1}^{N} F_j\Big)^2,
\end{equation}

where $F_i$ denotes the measured contact force along the constrained direction at the $i$-th sample within a sliding window of length $N$, and $\alpha$ is a scaling coefficient. 
This variance-based formulation captures the stability of the contact interaction: stable and consistent contact forces yield a small variance and thus a stronger constraint, while noisy or fluctuating forces produce a larger variance, which automatically weakens the influence of the contact factor and prevents the introduction of spurious constraints.

In the optimization, the adaptive weighting enters the objective through the Mahalanobis norm
\begin{equation}
    \| r_{\text{contact}}^{k}(\boldsymbol{\mathcal{X}}) \|^2 
    = \big(r_{\text{contact}}^{k}(\boldsymbol{\mathcal{X}})\big)^\top 
    \mathbf{P}_{\text{contact}}^{-1} 
    \, r_{\text{contact}}^{k}(\boldsymbol{\mathcal{X}}).
\end{equation}

\section{Hybrid Motion and Force Control with Image-based Visual Servo}
\label{sec: IBVS}

%\subsection{Hybrid Motion and Force Control}
%We exploit the capabilities of the fully-actuated UAV to design a hybrid motion-force controller. Unlike conventional coplanar multirotors, which can only produce thrust normal to the rotor plane and must tilt their entire body to align thrust with the desired force direction, a fully-actuated vehicle can regulate translation and orientation independently. 
Thanks to the fully-actuated property, we design the control input $\boldsymbol{\tau}^\mathcal{B}$ via feedback linearization as the combination of a motion controller $\boldsymbol{\tau}_{vs}^\mathcal{B}$ (driven primarily by visual servoing) and a wrench controller $\boldsymbol{\tau}_f^\mathcal{C}$ based on~\cite{guo2024flying, bodie2020active, he2023image}:
% \begin{align}
%     \boldsymbol{\tau} &= (\mathbf{I}_{6\times6}-\Lambda)\boldsymbol{\tau}_{vs} + \Lambda \boldsymbol{\tau}_f \label{eq:tau} \\ 
%     \boldsymbol{\Lambda} &= \text{blockdiag}(\boldsymbol{\Lambda}', \mathbf{0}_{3\times3}) \in \mathbb{R}^{6\times6} \label{eq:Lambda} \\ 
%     \boldsymbol{\Lambda}' &= {}^B_E\mathbf{R} \begin{bmatrix}
%         0 & 0 & 0 \\
%         0 & 0 & 0 \\
%         0 & 0 & \lambda(d) 
%     \end{bmatrix}, \label{eq:LambdaPrime}
% \end{align}
% where the selection matrix $\mathbf{\Lambda}$ specifies which DoFs are executed as direct wrench commands, while the complementary subspace is handled by motion control. The blending weight $\lambda(d) \in [0, 1]$ is determined by the camera-derived depth $d$ to the contact surface, smoothly transitioning from motion to force regulation as contact is approached. Because our end-effector is a rigid link, only the entry of $\mathbf{\Lambda}$ aligned with the pushing (surface-normal) direction is set to 1, while the remaining entries are all 0. Finally, the resulting command wrench $\boldsymbol{\tau}$ is mapped to rotor speeds by the PX4 controller via the standard allocation matrix.
\begin{equation}
    \bm{\tau}^\mathcal{B} = \mathbf{R}_\mathcal{B}^\mathcal{C}( ({\mathbf{I}_{6\times 6}}-\bm{\Lambda}){\mathbf{R}_{\mathcal{C}}^{\mathcal{B}}}\bm{\tau}_{vs}^\mathcal{B} + \bm{\Lambda} \bm{\tau}_f^\mathcal{C})
    \label{eq:tau}
\end{equation}
where $\bm{\Lambda} = \text{blockdiag}[\lambda(d), 0, 0,0,0,0] \in\mathbb{R}^{6\times6}, \lambda(d) \in \{0, 1\}$ is the selection matrix that specifies which DoFs are executed as direct wrench commands, while the complementary subspace is handled by motion control. $\lambda(d)$ is determined by the camera-derived depth $d$ to the contact surface, smoothly transitioning from motion to force regulation as contact is approached. $\lambda(d) = 0$ corresponds to pure motion control driven by visual servoing, while $\lambda(d) = 1$ activates force control along the surface normal direction.
%Specifically, $\lambda(d)$ in \eqref{eq:LambdaPrime} serves as the selected weight of hybrid control modes. $\lambda(d) = 0$ corresponds to pure motion control driven by visual servoing, while $\lambda(d) = 1$ activates force control along the surface normal direction. The wrench controller is therefore engaged once the aerial manipulator establishes contact. 
To enhance a smooth transition, we design $\lambda(d)$ as a confidence factor that increases progressively with the measured depth $d$:
\begin{equation}
    \label{eq:lambda}
    \lambda(d) = 
    \begin{cases}
        1, & d \leq d_{min}, \\
        \frac{1}{2}\left(1 + \cos \left(\frac{d-d_{min}}{d_{max}-d_{min}}\pi \right) \right), & d_{min} \leq d \leq d_{max}, \\
        0, & \text{otherwise}.
    \end{cases}
\end{equation}
%This formulation ensures a gradual and stable blending between motion and force control as the vehicle approaches the contact surface.
\vspace{-2em}

\subsection{Visual Servoing Strategy}
We use the standard image based visual servo~\cite{chaumette2006visual} with the velocity estimate from Sec.~\ref{sec: VIO} to guide the motion of the end-effector to align with the target on the surface.
%In this section, we illustrate a circle detection based IBVS to guide the motion of the end-effector towards the circular target on the wall.

Let us consider the real and desired image feature as $\bm{s}$, $\bm{s^*}$, respectively. The aim of IBVS is to minimize feature error $\bm{e}_s = \bm{s} - \bm{s^*}$. For typical aerial contact inspection with motionless target, $\bm{s^*}$ is constant, and changes in $\bm{s}$ depend only on camera motion. Then, the first-order feature error dynamics can be represented as:
\begin{align}
    \dot{\bm{e}}_s = \dot{\bm{s}} - \dot{\bm{s}}^* = \bm{L}(\bm{s})\bm{v}_c^\mathcal{O} -  \dot{\bm{s}}^* = \bm{L}(\bm{s})\bm{v}_c^\mathcal{O}
\end{align}

where $\bm{v}_c \in \mathbb{R}^{6}$ is the spatial velocity of the camera (camera twist), including the instantaneous linear and angular components. $\bm{L}(\bm{s}) \in \mathbb{R}^{k\times 6}$ ($k$ is the number of features) is the interaction matrix. We consider velocity input to the aerial manipulator controller. To impose an exponential decrease of the feature error, the camera twist can be designed as 
\begin{equation}
    \bm{v}_c^\mathcal{O} = - \zeta \hat{\bm{L}}(\bm{s})^\dagger \bm{e}_s
\end{equation}

where $\hat{\bm{L}}(\bm{s})^\dagger$ denotes the estimated Moore-Penrose pseudo-inverse of $\bm{L}(\bm{s})$. $\zeta > 0$ is the convergence rate. Then, the velocity input to the UAV is $\bm{v}^\mathcal{B}_{\text{des}} = \mathbf{R}_\mathcal{B}^\mathcal{O}\bm{v}_c^\mathcal{O}$, and a standard PID controller can be designed to obtain $\bm{\tau}_{vs}^\mathcal{B}$.

The key components are robust image–feature detection and accurate interaction matrix estimation. For the tasks such as peg-in-the hole, the target can be modeled as a circle. Standard circle detection and segmentation algorithm can be used. The detected image center $\bm{o}_c= [u, v]^\top$ and and the circle’s image area $A$ encode lateral alignment and range (scale), respectively: the alignment error measures the lateral offset between the end-effector tip and the target center, while the scaling error reflects distance to the target plane. We therefore choose the IBVS feature vector $\bm{s} = [\bm{o}_c^\top, r]^\top$, where $r$ is the radius in pixel. Estimated depth is given by $d = \sqrt{f_xf_y\frac{A_r}{A}}$ with $A$ the target area in pixels, \(A_r\) the known physical area of the circular target and \(f_x, f_y\) the camera focal lengths. $\bm{L}(\bm{s})$ can thus be obtained as:
\begin{align}
    \bm{L}(\bm{s}) &= \begin{bmatrix}
        -\frac{f_x}{d} & 0 & \frac{u}{d} & \frac{uv}{f_x} & -\frac{f_x^2 + u^2}{f_x} & v \\
        0 & -\frac{f_y}{d} & \frac{v}{d} & \frac{f_x^2 + v^2}{f_y} & -\frac{uv}{f_y} & -u \\
        0 & 0 & \frac{r}{d} & -\frac{rv}{f_y} & \frac{ru}{f_x} & 0
    \end{bmatrix}
\end{align}

\subsection{Wrench Controller}
We design an impedance controller to ensure reference force $F_r$ tracking of the end-effector during the physical interaction with the target. The onboard F/T sensor measures the interaction force $F$. Then, the controller is designed as:
\begin{align}
    e_f & = F - F_r, \quad
    e_{x}  = {\bm{e}_{s}}_{[3]} \notag\\
    F_f & = F_r - m m_{d}^{-1}(K_e e_{x} +  D_{e} {\dot{e}_{x}}) - K_{f,p}e_f -  K_{f,i}\int{e_f dt} \notag \\
    & = F_r - K_{e,p} e_{x} - K_{e,d} \dot{e}_{x} - K_{f,p}e_f -  K_{f,i}\int{e_f dt} 
    \label{eqn:impedance}
\end{align}
with $m$, $m_d$ the actual and desired vehicle mass, respectively, $K_{e,p}$, $K_{e,d}$ are the normalized stiffness, and damping, respectively. $K_{f,p}, K_{f,i} > 0$ are tunable gains for force control. ${\bm{e}_{s}}_{[3]}$ is the third component of $\bm{e}_s$, which reflecting the scaling error. Then, the control wrench $\bm{\tau}_f^\mathcal{C} = [F_f, \bm{0}_{1\times5}]^\top$ and $\bm{\tau}_{vs}^\mathcal{B}$ will be passed through \eqref{eq:tau} to get the total control wrench $\bm{\tau}^\mathcal{B}$, which is finally distributed to each rotor through the control allocation process.
Furthermore, leveraging the fully-actuated design, we command zero roll and pitch to keep the vehicle level during the manipulation task.

\section{Experiments}
\label{sec: experments}
We evaluate the proposed method in both simulation and real-world. In Gazebo simulation, we assess the robustness under varying levels of estimation error. Real-world experiments are then conducted to demonstrate contact-rich aerial manipulation using only onboard sensing.

\begin{figure*}[!t]
    \centering
    \includegraphics[width=1\textwidth]{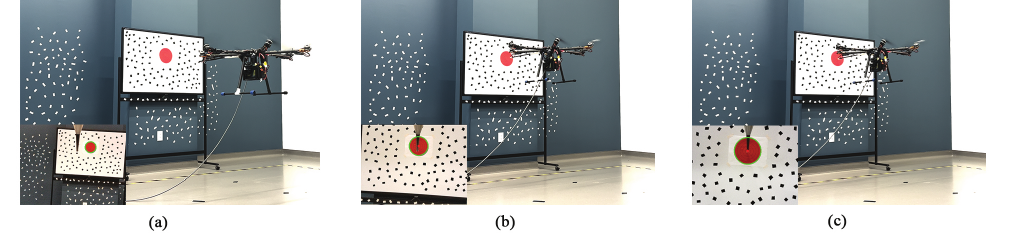}
    \caption{Snapshots of real-world experiments. (a) Visual servoing initialized. (b) Approaching the target. (c) Force holding.} 
    \label{fig: target_shooting}
    \vspace{-4mm}
\end{figure*}

\subsection{Experiment Setup}
In simulation, we model a fully actuated aerial manipulator in the Gazebo simulator following the UAV design in Sec.~\ref{subsec: systemDesign}. An RGB-D camera, configured with the same intrinsics/extrinsics as the VOXL2 Hi-Res optical sensor, is front-mounted at the same location on the vehicle. A 6-axis F/T sensor plugin is attached near the end-effector.

In real-world experiment, The aerial manipulator relies entirely on onboard sensing and computation as described in Sec.~\ref{subsec: methodOverview}. An OptiTrack motion-capture system provides ground-truth pose only for offline evaluation of the contact-aware VIO and is never used for feedback. 

%When IBVS is activated, the color-basd detection module will output segmentation results for the specific color, which are subsequently used to extract visual features. %Thus, we configure different target colors in both \ref{subsec: simResult} and \ref{subsec: realResult}.
% We consider a peg-in-hole task with force regulation as a representative aerial manipulation scenario of aerial contact inspection and maintenance. A circular target is mounted on a vertical surface. The aerial manipulator approaches the surface, performs insertion, and regulates a specified normal force while maintaining lateral alignment. 

We consider a peg-in-hole task with force regulation as a representative aerial manipulation scenario of aerial contact inspection and maintenance. The experimental procedure is as follows: First, the vehicle takes off and stabilizes in hover with the target in the camera’s field of view (FOV). Then the hybrid controller is activated and IBVS engaged. The color-based detector segments the target, and the resulting mask is used to extract visual features. Guided by our image-based visual servoing, the aerial manipulator approaches the surface, performs insertion, and regulates a specified normal force $F_d = 5$ N while maintaining lateral alignment. 

%a power line is connected externally to the force-torque sensor for power supply and communication purposes. We also pasted some tapes on the wall of the working area to create sufficient visual features for our VIO, as shown in Figure \ref{}.
\begin{figure}
    \centering
    \subfigure[Peg in Hole]{
        \includegraphics[width=0.22\textwidth]{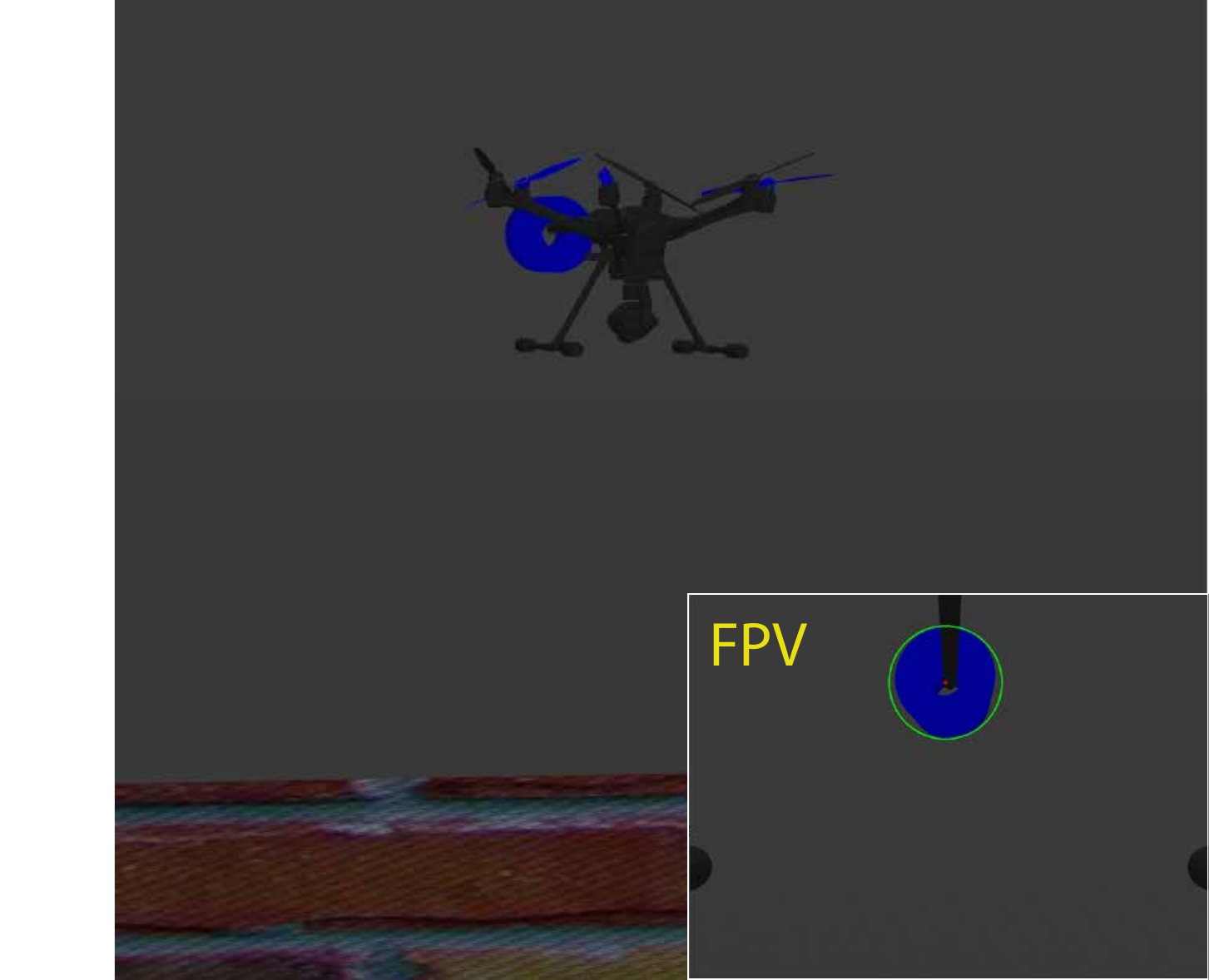}
    }\hspace{-2mm}
    \subfigure[Alignment Error]{
        \includegraphics[width=0.23\textwidth]{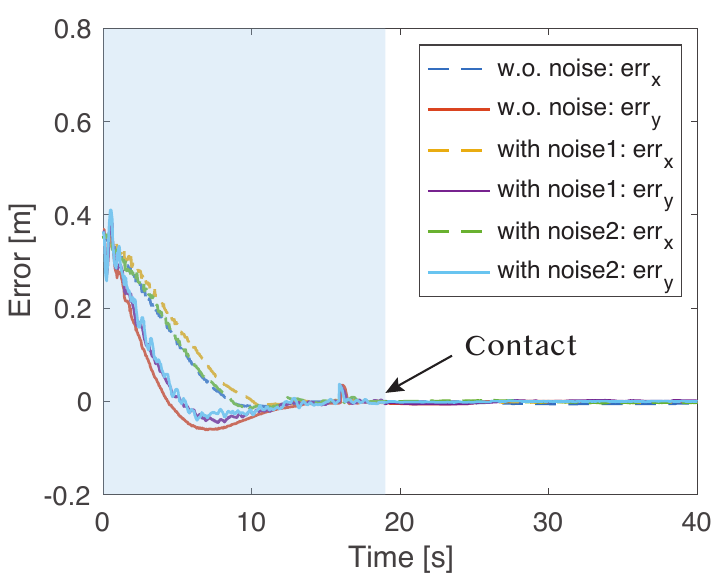}
    }\hspace{-2mm}\vspace{-2mm}
    
    \subfigure[Scaling Error]{
        \includegraphics[width=0.23\textwidth]{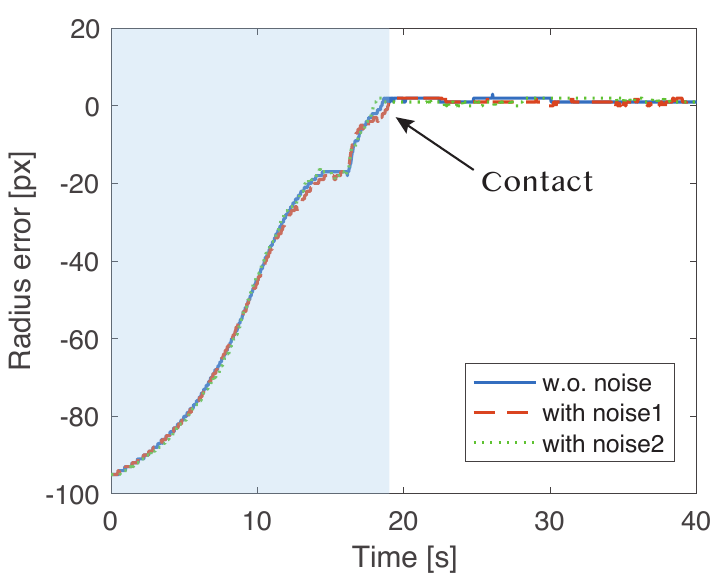}
    }\hspace{-2mm}
    \subfigure[Force Measurement]{
        \includegraphics[width=0.23\textwidth]{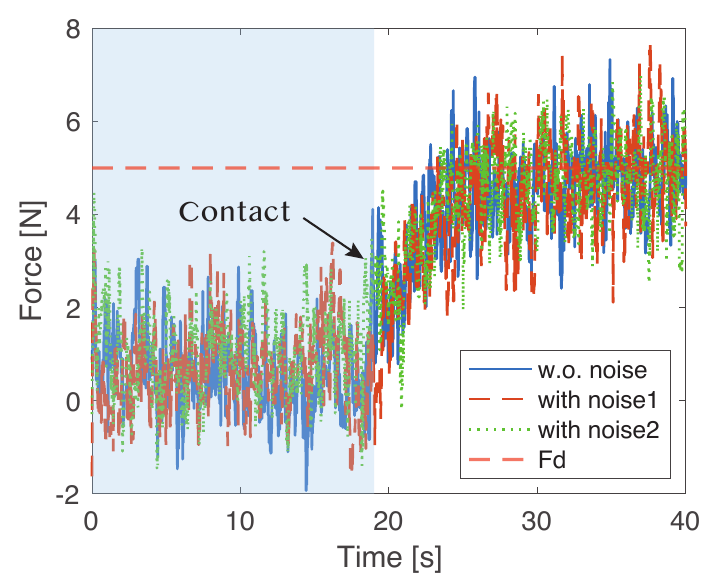}
    }\hspace{-2mm}
    \caption{Peg-in-hole experiment in simulation. (a) a snapshot of the contact moment. The first person view (FPV) is in the lower right corner. (b) Alignment error converted from the pixel plane. (c) Scaling error in pixel plane. (d) Filtered force measurement.}
    \label{fig: peginhole}
    \vspace{-4mm}
\end{figure}

\subsection{Simulation Results}
\label{subsec: simResult}
The experiment is conducted with a circular hole mounted on a vertical wall 1.5 m above the ground, with an inner diameter of 50 mm and an outer diameter of 140 mm; the aerial manipulator end-effector has a diameter of 12 mm. To verify the robustness of our proposed method, we repeat this experiment by gradually increasing levels of velocity estimation noise. The baseline uses ground-truth velocity.  In the second condition, we inject zero-mean, uniformly distributed noise with bounds ($\pm$ 0.05, $\pm$ 0.05, $\pm$ 0.07) m/s (noise 1). In the third, we increase the bounds to ($\pm$ 0.10, $\pm$ 0.10, $\pm$ 0.14) m/s (noise 2). Figure~\ref{fig: peginhole} summarizes force regulation and visual target tracking. In particular, Figures~\ref{fig: peginhole} (b)–(c) show IBVS feature tracking, where the lateral alignment error has been converted from pixels to meters using camera intrinsics and the target geometry, and the radius error is in pixel unit.  We can see despite the injected velocity errors, the image feature estimation still converge. The aerial manipulator successfully performs the peg-in-hole task and holds reference $F_d = 5$ N force. %Furthermore, under the target force of $F_d = 5$ N, the final force holding stage successfully demonstrates the expected performance.

\subsection{Real-world Experiments}
\label{subsec: realResult}

%In real-world environment, we demonstrate a similar target shooting with force regulation task. A printed red circular target is pasted on the whiteboard that firmly attached to the vertical wall. This target paper has a radius of 10 cm and contains four concentric circles at equal intervals. Following the similar procedure as in simulation experiments, the vehicle first takes off to a good viewpoint, then our hybrid control strategy will take over. Finally, the tip of end-effector will establish contact with the target center and maintain the given force.

\begin{figure}[htpb!]
    \centering
    \subfigure[Alignment Error]{
        \includegraphics[width=0.95\linewidth]{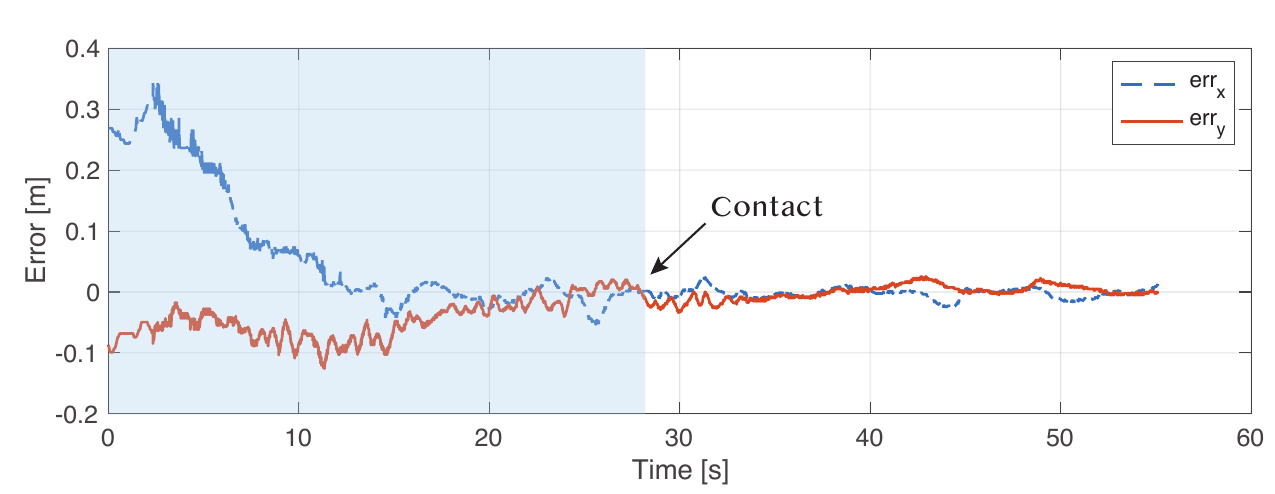}
    }\vspace{-5mm}
    \subfigure[Scaling Error]{
        \includegraphics[width=0.95\linewidth]{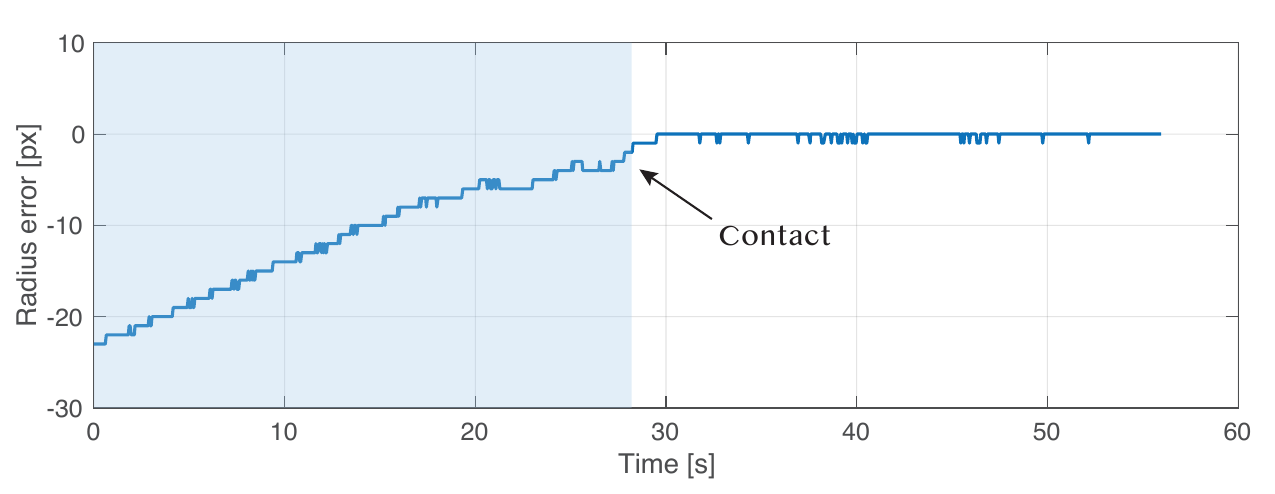}
    }\vspace{-5mm}
    \subfigure[Force Measurement]{
        \includegraphics[width=0.95\linewidth]{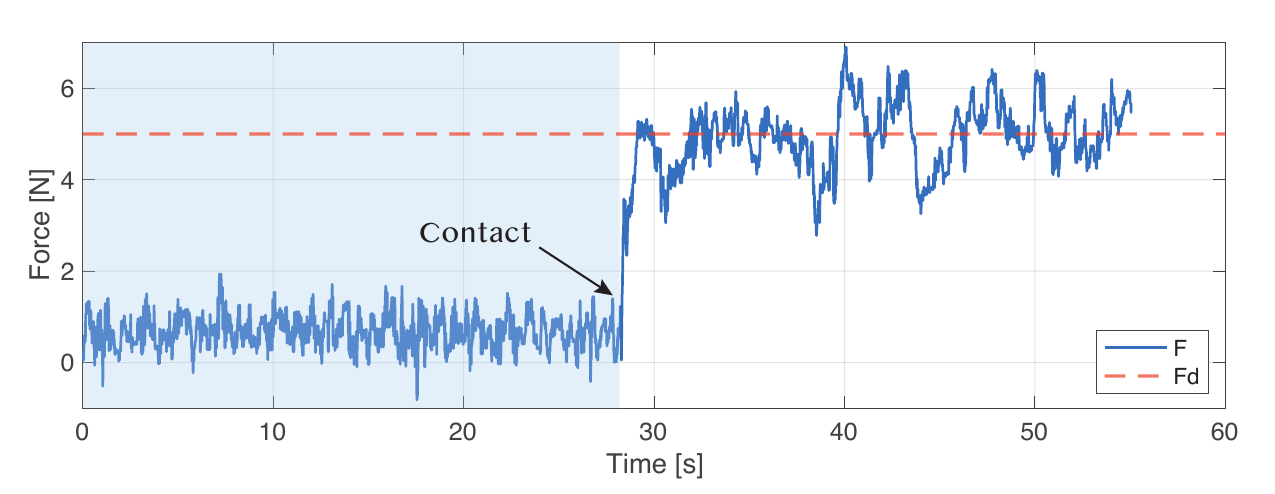}
    }\vspace{-5mm}
    \subfigure[Attitude]{
        \includegraphics[width=0.95\linewidth]{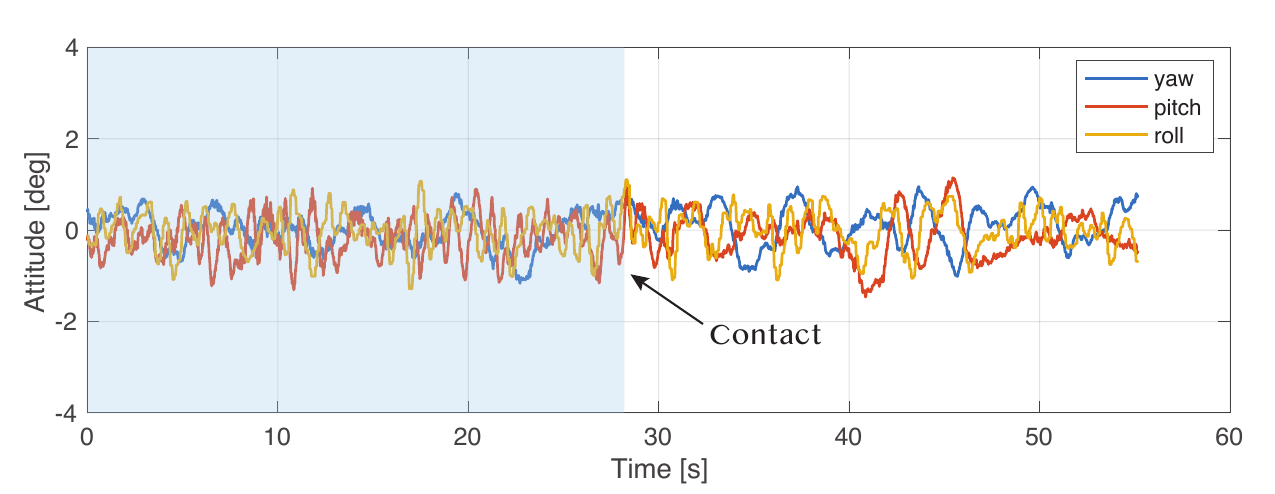}
    }\vspace{-1mm}
    \caption{Task evaluation in real world. (a) Alignment error converted from the pixel plane. (b) Scaling error in pixel plane (c) Filtered force measurement. (c) Flight attitude}
    \label{fig: target_shooting_plot}
    \vspace{-5mm}
\end{figure}

We then evaluate the task in real-world experiments. To assess robustness, we set up a red circular wall target with a radius of 10 cm and four equally spaced concentric rings. 
%To study the robustness of our method in real world, we choose to detect a red circular target on the wall, whose radius is 10 cm and containing four concentric circles at equal intervals. 
Figure.~\ref{fig: target_shooting} shows a sequence of flight snapshots (with an First person view (FPV) inset at the lower left). From \(t=0\) to \(28.3\,\mathrm{s}\), IBVS guides the vehicle toward the vicinity of target and the lateral alignment error converges to \(\approx 0\,\mathrm{m}\). At \(t=28.3\,\mathrm{s}\), the end-effector made contact with the target, indicated by a sharp rise in the measured force. From \(28.3\) to \(55.1\,\mathrm{s}\), the hybrid force–motion controller maintains the commanded normal force while simultaneously regulating lateral motion, as shown in Figure.~\ref{fig: target_shooting_plot} (a)–(b). Thanks to the fully-actuated mechanism, the platform also maintains a near-zero body attitude (roll/pitch) during contact (Figure.~\ref{fig: target_shooting_plot}(c)).%We further plotted the vehicle attitude during flight in Figure \ref{fig: target_shooting_plot} (c), to verify our hexarotor's capability of holding zero Euler angles at all times.

\begin{figure}[htpb!]
    \centering
    \subfigure[Velocity estimation in x-axis (contact direction)]{
        \includegraphics[width=1\linewidth]{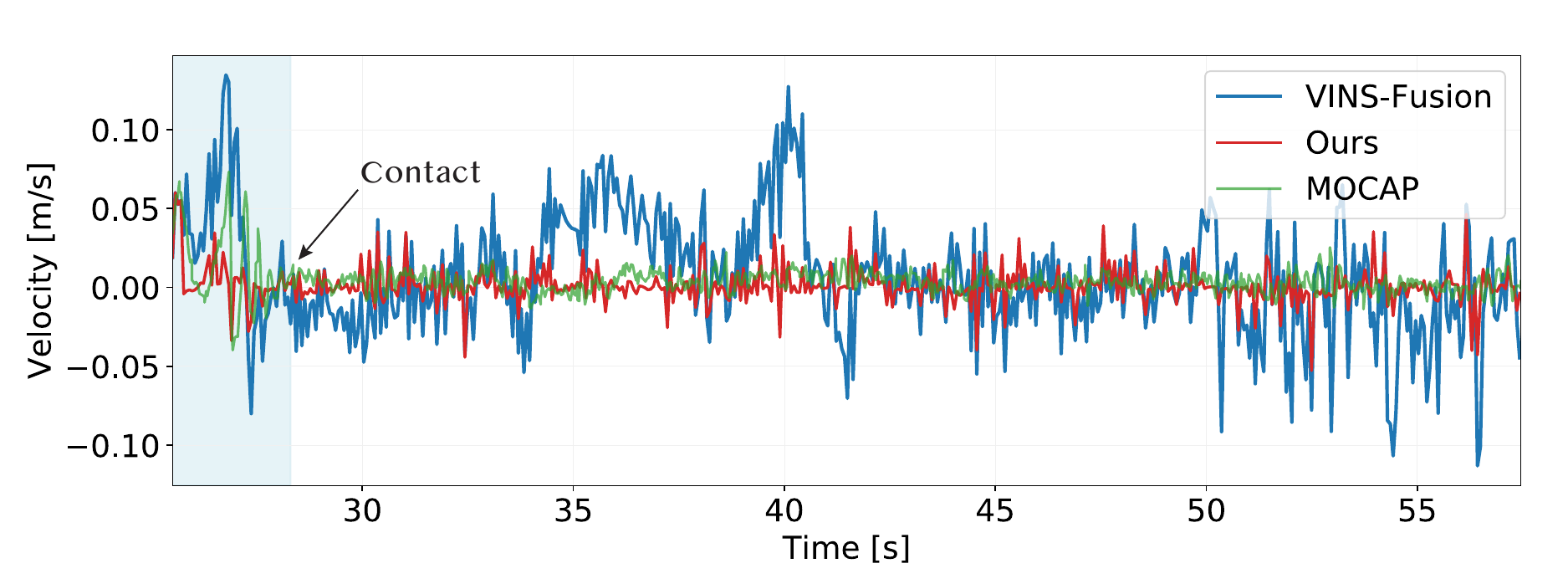}
    }\vspace{-3mm}
    \subfigure[Velocity estimation in y-axis]{
        \includegraphics[width=1\linewidth]{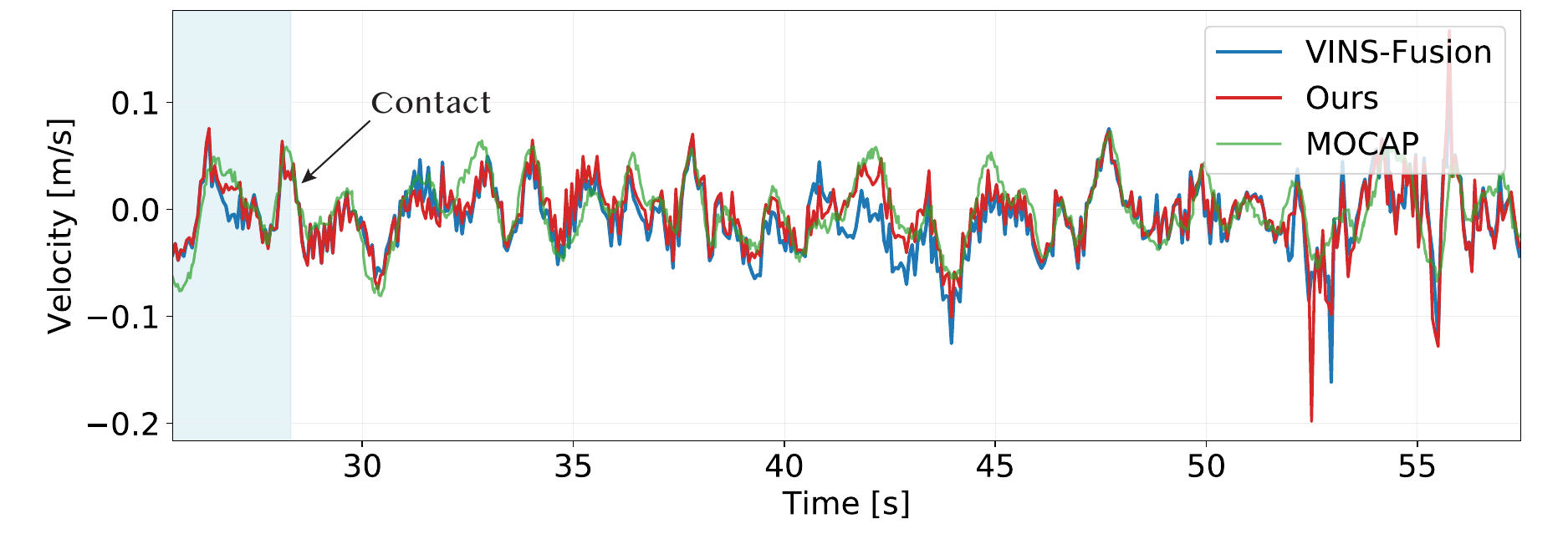}
    }\vspace{-3mm}
    \subfigure[Velocity estimation in z-axis]{
        \includegraphics[width=1\linewidth]{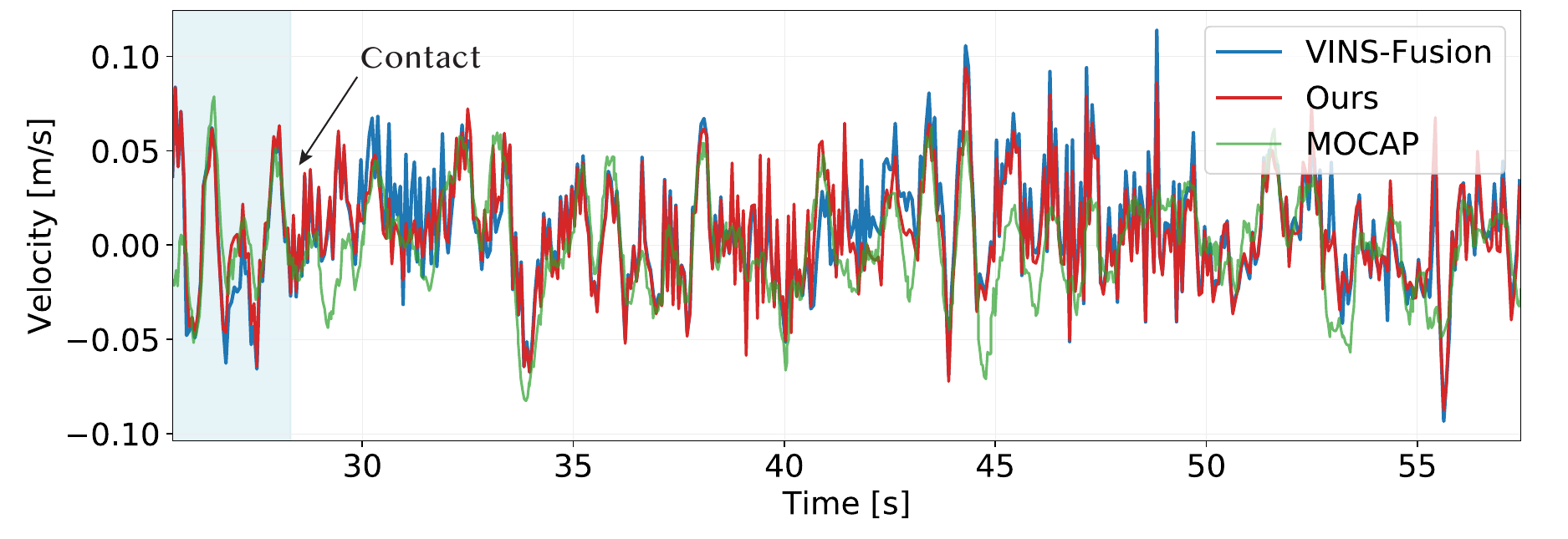}
    }\vspace{-3mm}
    \subfigure[VIO error in contact direction]{
        \includegraphics[width=1\linewidth]{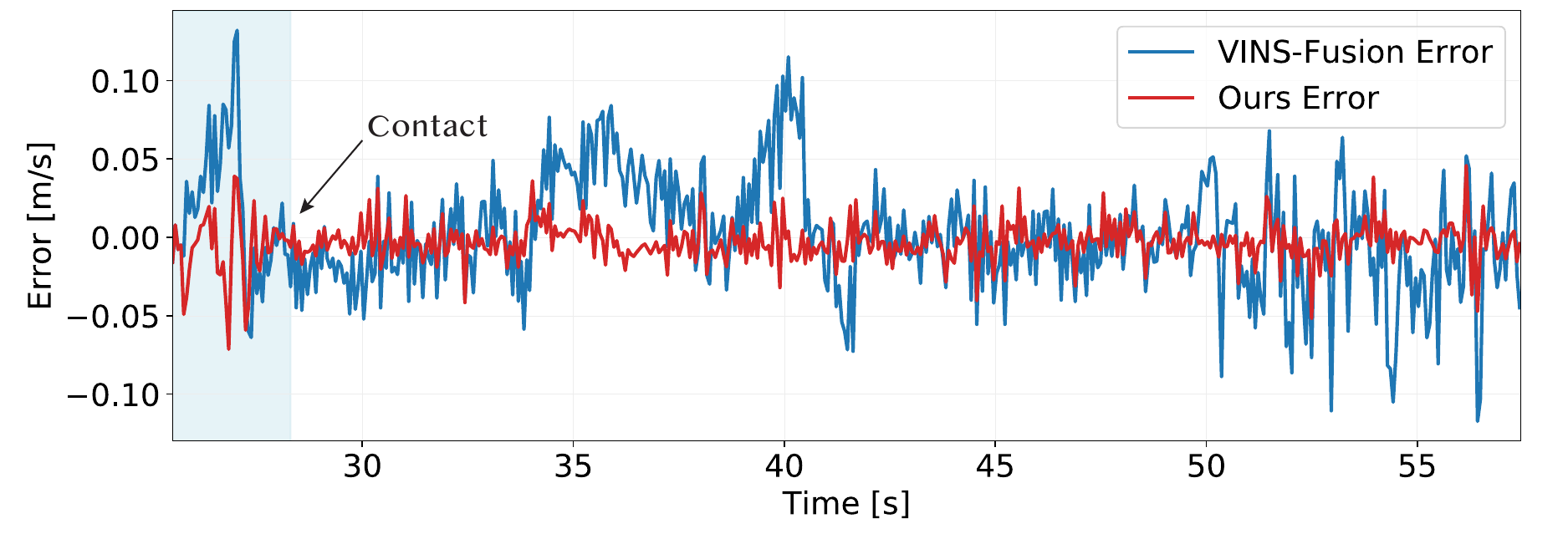}
    }\vspace{-2mm}
    \caption{Velocity estimation performance of VINS-Fusion and the proposed method, compared against ground-truth MoCap. (a)–(c) show velocity estimates along the $x$, $y$, and $z$ axes, respectively, with the $x$-axis corresponding to the contact direction. (d) presents the velocity estimation error along the contact axis}
    \label{fig: vio_error}
    \vspace{-5mm}
\end{figure}

% \begin{figure}[htpb!]
%     \centering
%     \includegraphics[width=0.48\textwidth]{Figure/Velocity_All_Axis.pdf}
%     \vspace{-2em}
%     \caption{VIO  estimation}
%     \label{fig:vio_error}
% \end{figure}

\begin{table}[h!]
    % \vspace{1.55mm}
    \centering
    \caption{Velocity Estimation Error during Contact (m/s) ($\downarrow$)}
    \vspace{-10pt}
    
    \begin{tabular}{c|cccc}
    \toprule
    Method & RMSE & mean & max & std   \\
    \midrule
    VINS-Fusion & 0.0356 & 0.0207 & 0.1174& 0.0356\\
    OpenVINS & 0.0619 & 0.0493 & 0.2123 & 0.0619 \\
    Ours & \textbf{0.0121} &\textbf{0.0089} & \textbf{0.0516} & \textbf{0.0117} \\
    \bottomrule
    \end{tabular}
    
    \label{table:VIO error}
    \vspace{-1.0em}
\end{table}

We also evaluate the state estimation performance of our contact-aware VIO by comparing it against VINS-Fusion \cite{qin2017vins} and OpenVINS \cite{geneva2020openvins}. 
%The experiments focus on the contact scenario, where the UAV makes physical interactions with the environment, which often introduces challenges for conventional VIO systems.
Figure~\ref{fig: vio_error} shows the estimated velocities along the $x$, $y$, and $z$ axes, 
together with the velocity error in the contact direction ($x$-axis). 
While all three estimators generally follow the ground truth motion-capture (MoCap) measurements, 
significant differences appear in the contact direction. 
In particular, VINS-Fusion exhibits noticeable fluctuations during contact, resulting in large estimation errors, whereas our method remains stable and closely aligned with the MoCap ground truth. The error plot further highlights that the proposed approach suppresses high-frequency noise and reduces drift along the contact axis, leading to smoother and more accurate estimates.

A quantitative comparison is provided in Table.~\ref{table:VIO error}, which reports the Root Mean Square Error (RMSE), mean, maximum, and standard deviation of the velocity error magnitude during contact. 
Our method achieves the lowest error across all metrics, with an RMSE of \textbf{0.0121~m/s} and mean error of only \textbf{0.0089~m/s}, significantly outperforming VINS-Fusion (RMSE 0.0356~m/s) and OpenVINS (RMSE 0.0619~m/s). 
These results clearly demonstrate that leveraging contact information substantially improves the robustness and accuracy of state estimation, particularly in challenging conditions where pure visual–inertial methods struggle.
\section{Conclusion}
\label{sec: conclusion}

% This paper developed a complete perception-control framework for aerial manipulation that eliminates reliance on an external motion capture system and enables robust contact-rich interactions. By incorporating contact-consistency factors into VIO, the estimator achieves improved accuracy during physical interaction, while the proposed IBVS and hybrid force-motion controller effectively regulate both contact wrenches and lateral motion. Experimental validation demonstrated accurate target approach, a 66.01\% reduction in velocity estimation drift at contact, and stable force holding, all achieved using only onboard sensing. These results highlight the potential of our pipeline for more complex outdoor aerial manipulation tasks. Future work includes online surface normal estimation, executing diverse aerial manipulation tasks under wind disturbance and in feature-sparse environments.

This paper developed a fully onboard perception–control pipeline for contact-rich aerial manipulation without external motion capture system. By adding contact-consistency factors to VIO and integrating IBVS with a hybrid force–motion controller, the system improves estimation during physical interaction and regulates both contact wrench and lateral motion. Experiments demonstrate accurate target approach, stable force holding, and a 66.01\% reduction in velocity estimation drift at contact with only onboard sensing, highlighting the potential for complex outdoor tasks. Future work will address online surface normal estimation and robustness to wind disturbance and feature-sparse environments.
\bibliographystyle{IEEEtran}
\bibliography{root}

\end{document}